# Synthetic Sampling for Multi-Class Malignancy Prediction

Matthew Yung, Eli T. Brown, Alexander Rasin, Jacob D. Furst, Daniela S. Raicu
Computing and Digital Media, DePaul University, Chicago, IL, 60604, USA, matt.yung23@gmail.com

**Abstract**

We explore several oversampling techniques for an imbalanced multi-label classification problem, a setting often encountered when developing models for Computer-Aided Diagnosis (CADx) systems. While most CADx systems aim to optimize classifiers for overall accuracy without considering the relative distribution of each class, we look into using synthetic sampling to increase per-class performance when predicting the degree of malignancy. Using low-level image features and a random forest classifier, we show that using synthetic oversampling techniques increases the sensitivity of the minority classes by an average of 7.22% points, with as much as a 19.88% point increase in sensitivity for a particular minority class. Furthermore, the analysis of low-level image feature distributions for the synthetic nodules reveals that these nodules can provide insights on how to preprocess image data for better classification performance or how to supplement the original datasets when more data acquisition is feasible.

## 1 INTRODUCTION

In the radiology domain, radiologists interpret, characterize, and make inferences from data acquired with various medical imaging modalities. However, human interpretations are subjective and can often lead to diagnostic errors [1]. Computer-Aided Diagnostics (CADx) systems have been proposed as "second readers" to augment and improve the medical image interpretation process.

When diagnostic ground truth is not available, a reference truth formed from human interpretations is used to derive the class labels. However, given the complexity and difficulty of the diagnostic interpretation process, it can be challenging to provide a deterministic label. This complication is present in most biological systems in which there is gradation between normal and abnormal [2]. Furthermore, setting thresholds for transforming the problem to the two conventional states, malignant versus benign, is difficult as thresholds can vary among human observers. Therefore, it is necessary to investigate the performance of CADx systems in the context of multi-class problems in which the diagnostic classes represent degrees of uncertainty such as: 1= 'highly unlikely', 2 = 'moderately unlikely', 3 = 'indeterminate', 4 = 'moderately suspicious', and 5 = 'highly suspicious'. Including the degree of uncertainty ('lack of diagnostic confidence') into CADx performance evaluation will certainly address the deterministic output limitation of most current CADx systems.

This variability in experts' interpretation introduces a level of uncertainty to the labels. In addition, lesions of "unique cases" are rare, which creates a class imbalance [3]. When the imbalance is severe, predictive models suffer in generalizability by biasing towards the majority class. In order to facilitate the development of improving such systems, the NIH/NCI Lung Image Database Consortium (LIDC) was introduced [4]. The LIDC data, a collection of thoracic CT scans with detected pulmonary nodules annotated by four expert radiologists, serves as an example of the many difficulties encountered when analyzing medical data. With no clear ground truth and wide variability in agreement among the radiologists, the annotated malignancy labels introduce a high degree of uncertainty into the CADx process. The ratings of nodule characteristics are also on a multi-rated scale which exacerbates class imbalance concerns. Since minority classes are often the more "interesting" cases, it is important to classify these cases accurately.

In this paper, rather than studying a benign/malignant classification problem, we consider all five class ratings of malignancy. Given the degree of uncertainty for malignancy, studying the multi-class classification problem has to be augmented with solving the class imbalance problem. In order to address class imbalance, sampling techniques to rebalance the dataset for training are typically used. One approach is to oversample the minority classes to match the number of instances in the majority class. A popular algorithm is the Synthetic Minority Over-sampling Technique (SMOTE) [5], which synthetically creates new instances. However, blindly using an oversampling technique can lead to unexpected results. Therefore, we will investigate the effects of four well-known oversampling techniques, all of which generate new samples synthetically, on the multi-classification problem. Moreover, we will provide an analysis of the synthetic sampling techniques by comparing individual low-level image feature's distribution for each oversampled dataset against the original dataset. The main contribution of this work is that this study is the first for the LIDC dataset, as far as we know. Furthermore, this analysis will provide general insights on the types of data generated as well as how it affects the classifier when using oversampling techniques. The overall methodology is shown in Fig. 1. Although the analysis is presented in the context of the LIDC dataset, the same approach can be applied for other datasets that exhibit problems associated with class imbalance.

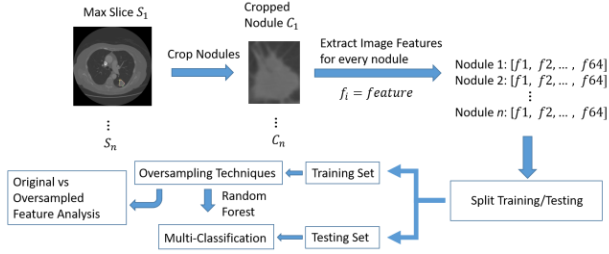

**Figure 1: Diagram of the proposed methodology**

## 2 Related work

Since the LIDC has been made publically available in 2012, there have been many studies with the intent of improving classification of the semantic characteristics, especially malignancy, provided in the LIDC dataset. For instance, Vinay et al. [6] compared homogenous versus heterogeneous ensemble of classifiers for the prediction of malignancy refined to a binary classification problem using 54 extracted low-level image features. They used the LIDC semantic characteristics to create a binary class conversion rule. Kumar et al. [7] used a binary tree malignancy classifier using 200 features learned from an auto-encoder neural network and the biopsy, surgical resection, or progression available for a small subset of cases.

Moving forwards from a binary classification, a few approaches have been developed for a multi-label classification. Instead of predicting the actual class rating directly, Zinovev et al. [8][9] proposed belief decision trees and ensemble of classifiers with active DECORATE learning to predict probabilities for the class ratings. Kaya et al. [10] presented an approach using an ensemble classifier and a rule based fuzzy inference method. Given that the LIDC data, like many other medical datasets, is affected by problems related to unbalanced datasets, both studies also evaluated their classifier on balanced dataset by performing under-sampling [8] and over-sampling using SMOTE [10].

An analysis across different synthetic oversampling algorithms was conducted by Abdi et al. [11] in which the authors proposed a new over-sampling algorithm based on Mahalanobis distance and compared the effects on a multi-class dataset. They showed that their proposed method ultimately generates less duplicated and overlapped data points as opposed to all other oversampling algorithms. In a similar manner, we propose to compare four oversampling techniques to determine the effects of oversampling for a multi-classification problem on a feature level rather than observation level in the context of LIDC.

## 3 Methods

### 3.1 The LIDC Data and low-level image features

The LIDC dataset currently contains CT scans for 1010 patients, each averaging about 200-300 CT associated images [5]. Nodules were rated by four radiological experts across nine semantic characteristics (calcification, internal structure, lobulation, malignancy, margin, sphericity, spiculation, subtlety, and texture) on a scale from one to five.

All four radiologists did not need to come to a consensus when detecting and rating a nodule among the total of 2,687 identified nodules. Therefore, each nodule could be rated by one, two, three, or all four radiologists. Since the focus of this paper is not on the detection and diagnosis variability, we only consider 829 nodules that are rated by all four radiologists and encode the variability of the experts' malignancy ratings using the mode of their ratings. If no unique mode was determined, the ceiling of the mean was used. This resulted in 829 unique nodule instances, each with an associated aggregate malignancy rating.

Following image extraction and preprocessing procedures as shown in previous works [12][8], each nodule instance is represented by the slice containing the largest region of interest based on area in pixels amongst the four radiologists' marked spatial boundaries. We extracted 65 low-level image features (intensity, shape, size, and texture) from each nodule greater than an area of 25 pixels.

### 3.2 Oversampling

One of the main issues in LIDC is the unbalanced semantic characteristic distribution. Of the 829 unique instances, 351 (42.34%) belong to malignancy class 3 ('indeterminate') and the rest are in four different classes (Table 1). To account for the unbalanced class distribution, we explore four oversampling techniques: SMOTE [5], SMOTE Borderline 1 (SMOTE B1) and Borderline 2 (SMOTE B2) [13], and adaptive synthetic sampling approach (ADASYN) [14]. These four approaches synthetically generate new minority samples from K-Nearest Neighbors (K-NN). In addition to the four explored oversampling approaches, random oversampling was applied for baseline comparison. Before discussing the procedural differences between SMOTE and its variants, we define some terms below.

The minority class and its set of instances are defined as $\{s_1^{min}, s_2^{min}, \ldots, s_N^{min}\}$ where $min = \overline{1, L}$ and $min \neq 3$ due to being a majority class, and $N = \#instances$ in class $min$. Each minority instance is represented as $s_j^{min} = [f_{j1}^{min}, f_{j2}^{min}, \ldots, f_{jp}^{min}]$ for $j = \overline{1, N}$ and the number of features $p = \overline{1, 65}$. Additionally, each minority instance is associated with a set of K-NNs $s_j^{min} = [a_{j1}^r, a_{j2}^r, \ldots, a_{jK}^r]$; each neighbor is represented as $a_{jo}^r = [fa_{jo,1}^r, fa_{jo,2}^r, \ldots, fa_{jo,p}^r]$ where $r \in L$ and $o = \overline{1, K}$. We denote the number of out-of-class neighbors as $H$. Oversampled instances are denoted as $syn_q^{min} = [fsyn_{q1}^{min}, fsyn_{q2}^{min}, \ldots, fsyn_{qp}^{min}]$ where $q$ is the number of synthetic instances to generate.

**Table 1: Class distribution amongst the 829 dataset.**

|  | Class 1 | Class 2 | Class 3 | Class 4 | Class 5 |
|---|---|---|---|---|---|
| **Total Frequency** | 117 | 85 | 351 | 166 | 110 |
| **Total Percentage** | 14.11% | 10.25% | 42.34% | 20.02% | 13.27% |

SMOTE and its variants generate new synthetic instances by interpolating new feature values between the minority instance and its respective nearest neighbor. SMOTE only considers within class neighbors when randomly generating more synthetic minority samples. SMOTE B1 and SMOTE B2 generate synthetic samples on instances considered in "DANGER," determined when most of their neighbors belong to the out-of-class instances. SMOTE uniformly interpolates new feature values between a minority instance and its respective randomly selected neighbors, independently on each feature equation (1).

$$fsyn_{qp}^{min} = f_{jp}^{min} + \lambda(fa_{jo,p}^{L} - f_{jp}^{min}), \lambda = rand[0,1] \quad (1)$$

In SMOTE B1, the interpolation process for generating new synthetic minority samples is the same as original SMOTE, applied only on the instances assigned as in "DANGER," a subset of the minority class. Each minority instance is assigned a label (Safe, Danger, or Noise) based on a criterion from its respective $K$ neighbors equation (2). SMOTE B2 follows a similar procedure as SMOTE B1, with the only difference being that SMOTE B2 generates new synthetic instances calculated on all neighbors instead of only within-class neighbors.

$$Assign(s_j^{min}) = \begin{cases} Safe, 0 \le \frac{K}{2} \\ Danger, \frac{K}{2} \le H < K \\ Noise, \ H = K \end{cases} \quad (2)$$

ADASYN synthetically generates new minority samples using a density distribution based on the number of out-of-class neighbors. A minority instance surrounded by more out-of-class instances is considered hard-to-train, and is thus given a higher probability to be augmented by generating synthetic samples. Ultimately, ADASYN is a pseudo-probabilistic algorithm in the sense that a fixed number instances is generated for each minority instance based on a weighted distribution of its neighbors.

### 3.3 Jensen-Shannon Divergence

A visual and quantitative analysis was further conducted to compare feature distributions between the original training data and each of the oversampled training data. In particular, we analyzed whether the synthetic samples generated through each algorithm still produced a dataset representative of the original set. For each of the 65 image features, we plotted the frequency distributions for each class between the original training data and the oversampled training data. The differences between the distributions were quantified and compared using the Jensen-Shannon (JS) similarity, the square root of JS divergence [15].

## 4 Results

### 4.1 Classification

For this study, a random forest classifier with 30 iterations of stratified repeated hold-out partitioning technique was used. The training data was oversampled such that all the number of cases in the minority classes (1, 2, 4, 5) was the same number of cases as the majority class (class 3).

Classifiers on highly unbalanced datasets usually produce biased predictions for the majority case. As expected, Table 2 shows that, without any oversampling, majority class 3 ('indeterminate') had the highest sensitivity. Surprisingly, no correct prediction was made for class 2 ('moderately unlikely').

There are obvious tradeoffs from balancing the training set. In particular, overall classification accuracy decreases by about 2 – 6%, as shown in Table 3, and class 3 sensitivity decreased by up to ~20%. When the classifier was trained with a balanced dataset through synthetically oversampling, sensitivity significantly improved. For instance, class 2 had a 0.57% sensitivity rate trained without any oversampling. In particular, SMOTE and its variants increased the sensitivity rate of class 2 by ~19% points. Random oversampling was outperformed by all the synthetic generating sampling techniques other than for class 4 against ADASYN in terms of sensitivity. Even so, ADASYN managed to outperform all other techniques for class 1 and class 5.

**Table 2: Multi-class sensitivity results from test set; average increase is based only on the synthetic oversampling algorithms (excludes random sampling)**

| Sensitivity | | No Oversampling | Oversampling | | | | | Avg. % Increase |
|---|---|---|---|---|---|---|---|---|
| | | | SMOTE | B1 | B2 | ADASYN | Rand | |
| Malignancy | 1 | 80.97 | 85.25 | 85.42 | 86.01 | **88.98** | 82.98 | 5.45 |
| | 2 | 0.57 | **19.88** | 18.48 | 19.28 | 15.49 | 7.63 | 17.71 |
| | 3 | **83.3** | 67.42 | 66.99 | 67.62 | 63.76 | 79.31 | -16.85 |
| | 4 | 53.17 | **59.58** | 58.93 | 54.12 | 47.07 | 53.77 | 1.76 |
| | 5 | 67.99 | 71.63 | 67.97 | 71.59 | **76.54** | 67.64 | 3.94 |

**Table 3: Overall accuracy results from test set**

| | No Oversampling | Oversampling | | | | |
|---|---|---|---|---|---|---|
| | | SMOTE | B1 | B2 | ADASYN | Rand |
| Overall Accuracy | **66.36** | 64.03 | 62.97 | 62.85 | 60.70 | 66.01 |

### 4.2 Oversampling Analysis

The quantitative analysis showed that generally the same types of features appear as the most different in its respective oversampling techniques between the oversampled training set and the original training set. Across each of the four oversampling techniques, independently ranked in descending order of JS similarity, class 1 and class 2's top features were texture features. On the other hand, class 4 and class 5's top features were intensity features. Also interesting to note that, on average, ADASYN produced higher JS similarity by as much as about 40% from the next leading oversampling technique (SMOTE B2), as shown in Table 4. In fact, one of the *Gabor* features produced a JS similarity as high as 0.96 under ADASYN. This result is due to the nature of the algorithm, and is further supported by visualizing the feature distributions in Fig. 2 and Fig. 3. The frequency distributions for each image feature were plotted for the original training set and for each of the oversampled training set at the class level. We selected two features of interest based on the ranked JS similarity values, an intensity feature and a texture feature showing the differences for class 4 and class 2, respectively. Although the differences in the frequency distributions did vary across classes and features, the selected features are chosen primarily to reflect the effects of oversampling rather than showing clear differences amongst classes. From Fig. 2, the original distribution for *MinIntensity* shows a major peak ($x = 125$) and a minor peak ($x = -800$) with no values between. All four synthetic generating algorithms, however, produced values between the two peaks. This is due to the nature of the interpolation process, which is a uniform shift on random neighbors calculated with Euclidean distance.

**Table 4: Average JS Similarity per oversampling technique**

| Average JS Similarity | | Oversampling | | | |
|---|---|---|---|---|---|
| | | SMOTE | B1 | B2 | ADASYN |
| Class | 1 | **0.1625** | 0.1847 | 0.2120 | 0.3367 |
| | 2 | **0.2078** | 0.2223 | 0.2578 | 0.3713 |
| | 4 | 0.1416 | **0.1368** | 0.2015 | 0.2523 |
| | 5 | **0.1549** | 0.1587 | 0.2059 | 0.2996 |
| Overall | | **0.1667** | 0.1756 | 0.2193 | .3150 |

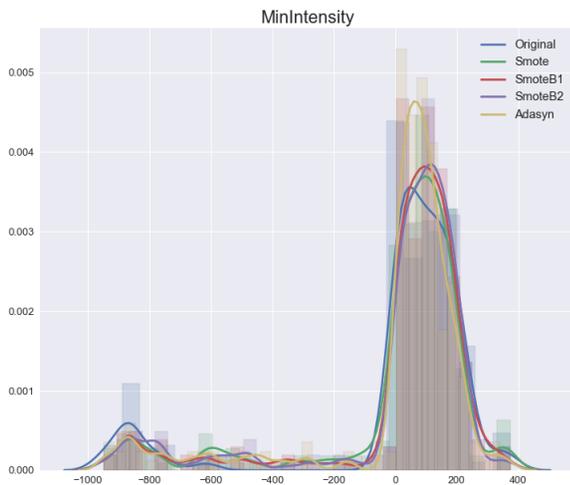

**Figure 2: Frequency distribution of selected intensity feature**

The idea of ADASYN's algorithm was to create synthetic instances based on harder-to-train cases derived from a weighted probability of majority neighbors, which is the main reason for ADASYN's higher JS similarity compared to the other. The texture distribution explains just that in Figure 3. For instance, in the plot for *GaborSD_0_2*, there are indeed outliers in the original distribution in the range between $x = 0$ to $x=50$, where we don't actually see the curve. On the other hand, we start to see a curve for ADASYN's distribution (very small peaks of yellow within the same range). In ADASYN, some of the harder-to-train cases are the outliers, thus the algorithm produces some synthetic instances based on the outliers making it sensitive to outliers. The SMOTE algorithms are less sensitive to outliers, but not robust enough because theoretically there is a chance for a neighbor to be an outlier and to be randomly selected to generate synthetic instances. However, ADASYN adds a higher weight to such cases, increasing the chances to generate instances based on outliers.

This type of oversampling behavior could be problematic for classifiers if these synthetic values are just representatives of noise or outliers. In other words, the blind use of oversampling techniques has the potential to introduce "false" artificial samples of the original dataset. This indicates that, especially in the medical domain, careful attention should be given not only to improving CAD classifiers' performance, but also to generating data used for training. Therefore, better approaches to synthetically create instances that mimic the original distribution of the feature data have to be developed.

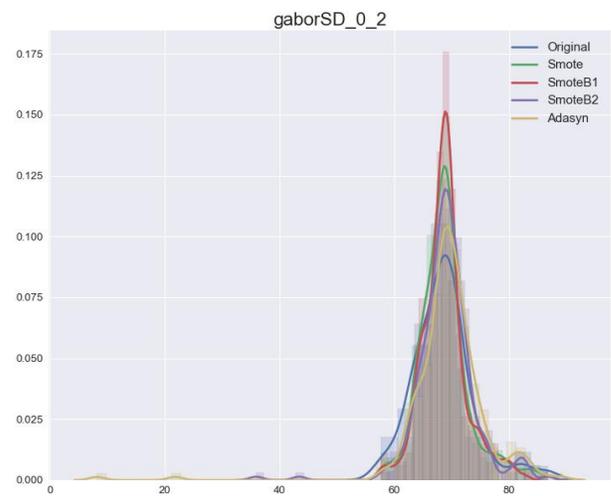

**Figure 3: Frequency distribution of texture features**

## 5 Conclusion

Our results show that by synthetically generating new minority cases, sensitivity of the minority classes are improved; for example, sensitivity improves significantly from 0.57% to as much as 19.88% for malignancy class 2 ('moderately unlikely'). For all minority class ratings, the synthetic oversampling algorithms gave the best performance in term of sensitivity (an improvement average of 7.22% points), but did also lower the overall accuracy because of the decrease in sensitivity for the majority class. Therefore, based on the discussed results, it is important to take notice of the performance tradeoffs between minority and majority classes when balancing the datasets.